\documentclass[runningheads]{llncs}

\usepackage[T1]{fontenc}
\usepackage{amsmath,amssymb,amsfonts}
\usepackage{makecell}
\usepackage{threeparttable}
\usepackage{algorithmic}
\usepackage{multicol}
\usepackage{multirow}
\usepackage{bbding}
\usepackage[misc]{ifsym}
\usepackage{xcolor}
\usepackage{graphicx,verbatim}
\usepackage{booktabs}

\begin{document}

\title{InViC: Intent-aware Visual Cues for Medical Visual Question Answering}
\titlerunning{InViC for Med-VQA}

\author{Zhisong Wang$^{1}$ \and
Ziyang Chen$^{1}$ \and
Zanting Ye$^{3}$ \and
Hongze Zhu$^{2}$ \and
Yefeng Zheng$^{2}$$^{(\textrm{\Letter})}$ \and
Yong Xia$^{1}$$^{(\textrm{\Letter})}$}  
\authorrunning{Z. Wang et al.}

\institute{
National Engineering Laboratory for Integrated Aero-Space-Ground-Ocean Big Data Application Technology, School of Computer Science and Engineering, Northwestern Polytechnical University, Xi’an 710072, China\\
\and
Westlake University \\
\and
Southern Medical University \\
\email{zswang@mail.nwpu.edu.cn}\\
}
  
\maketitle    

\begin{abstract}

    Medical visual question answering (Med-VQA) aims to answer clinically relevant questions grounded in medical images. However, existing multimodal large language models (MLLMs) often exhibit \emph{shortcut answering}, producing plausible responses by exploiting language priors or dataset biases while insufficiently attending to visual evidence. This behavior undermines clinical reliability, especially when subtle imaging findings are decisive. 
    We propose a lightweight plug-in framework, termed \textbf{Intent-aware Visual Cues (InViC)}, to explicitly enhance image-based answer generation in medical VQA. InViC introduces a \emph{Cue Tokens Extraction (CTE)} module that distills dense visual tokens into a compact set of $K$ question-conditioned cue tokens, which serve as structured visual intermediaries injected into the LLM decoder to promote intent-aligned visual evidence. To discourage bypassing of visual information, we further design a two-stage fine-tuning strategy with a cue-bottleneck attention mask. In Stage~I, we employ an attention mask to block the LLM's direct view of raw visual features, thereby funneling all visual evidence through the cue pathway. In Stage~II, standard causal attention is restored to train the LLM to jointly exploit the visual and cue tokens.
    We evaluate InViC on three public Med-VQA benchmarks (VQA-RAD, SLAKE, and ImageCLEF VQA-Med 2019) across multiple representative MLLMs. InViC consistently improves over zero-shot inference and standard LoRA fine-tuning, demonstrating that intent-aware visual cues with bottlenecked training is a practical and effective strategy for improving trustworthy Med-VQA.
    
\keywords{Med-VQA \and Intent-aware Visual Cues \and Visual Perception.}
\end{abstract}

\section{Introduction}

Medical visual question answering (Med-VQA) extends visual question answering to clinical imaging, requiring models to answer natural-language questions about anatomy, findings, and abnormalities conditioned on medical images (e.g., radiography, CT, or MRI)~\cite{lin2023medical}. By enabling interactive, image-grounded querying, Med-VQA holds significant promise for clinical decision support, report assistance, and medical education. Recent progress has been largely driven by multimodal large language models (MLLMs) and large-scale medical vision-language pretraining. Injecting domain-specific visual knowledge into MLLMs improves multimodal reasoning~\cite{chen2024towards}, while medical foundation models aim to unify perception and reasoning across tasks and modalities~\cite{xu2025lingshu}. Stronger LLM backbones enhance instruction following and biomedical knowledge coverage, improving robustness to heterogeneous phrasing and cross-dataset transfer.

Despite these advances, a critical reliability issue persists: \emph{shortcut answering}. Models may exploit language priors, recurring question templates, or dataset biases to produce plausible answers without adequately grounding them in the image. Such behavior is especially problematic in clinical scenarios, where subtle visual cues can be diagnostically decisive and plausible-but-incorrect responses may mislead downstream decisions.
Existing mitigation strategies address shortcut answering from several complementary perspectives.
\textbf{(1) External knowledge and agentic reasoning.}
Some methods augment MLLMs with additional clinical knowledge to reduce question-driven guessing. Retrieval-augmented generation~\cite{yuan2023ramm} and knowledge-graph frameworks~\cite{wu2025mkgf} incorporate structured or retrieved medical information to support answer generation. Agentic systems further decompose reasoning into planning, verification, or tool-use stages, including dynamic agent frameworks~\cite{xiao2025dynamic} and multimodal medical agents equipped with tools~\cite{li2024mmedagent}.
\textbf{(2) Bias-aware training objectives.}
Another line of work explicitly targets textual bias through data design and objective-level interventions. DeCoCT mitigates terminology-driven shortcuts via counterfactual question construction and contrastive learning~\cite{wan2025eliminating}, while Med-BiasX introduces a bias-aware framework with a question-only branch and calibration objectives to suppress reliance on language priors~\cite{zhu2025med}.
\textbf{(3) Alignment between pretraining and Med-VQA supervision.}
Other efforts aim to reduce the semantic gap between large-scale pretraining and downstream Med-VQA tasks. Prompt learning alleviates distribution mismatch without major architectural changes~\cite{lu2025bridging}. Latent prompt assistance generates answer-conditioned prompts to extract relevant evidence~\cite{gu2024lapa}, and difference VQA with residual alignment emphasizes clinically meaningful changes in longitudinal settings~\cite{lu2024spot}.

Despite this progress, a fundamental limitation remains: the \emph{internal evidence pathway} from image representation to answer generation can still be bypassed. In current MLLM architectures, dense visual tokens are typically concatenated with textual tokens without structural constraints. Under modality imbalance and strong language priors, the decoder may prioritize question cues over image features, especially when visual evidence is subtle or dataset shortcuts are salient, leading to plausible but weakly grounded answers. 
Although debiasing objectives can attenuate shortcut reliance~\cite{wan2025eliminating,zhu2025med} and prompting strategies can improve semantic alignment~\cite{lu2025bridging}, these approaches generally do not explicitly regulate how visual information must enter and influence the generation process. For clinically credible Med-VQA, correct predictions should consistently arise from image-grounded evidence rather than question priors alone.

In this work, we argue that Med-VQA requires not only better objectives or prompts, but also an explicit architectural constraint on how visual information is consumed. To this end, we propose a lightweight plug-in framework, termed \textbf{Intent-aware Visual Cues (InViC)}, to explicitly enhance image-based answer generation in medical VQA.
InViC introduces a \emph{Cue Tokens Extraction (CTE)} module that distills dense visual tokens into $K$ \emph{intent-aware cue tokens}. These cues are initialized from a pooled question representation and refined via cross-attention to question and visual tokens, forming a compact, task-focused summary of image evidence. The cues are then injected into the LLM decoding context as structured visual intermediaries. 
Crucially, we introduce a two-stage training strategy. In Stage~I, we employ a cue-bottleneck attention mask to block the LLM's direct view of raw visual tokens, forcing image evidence to flow exclusively through cue tokens. In Stage~II, we restore standard causal attention to match inference conditions while retaining the learned cue pathway. This visibility scheduling encourages the model to internalize cue-mediated grounding without permanently restricting expressivity.
We evaluate InViC on three public Med-VQA benchmarks across multiple representative MLLMs. Our results demonstrate consistent improvements over zero-shot and LoRA baselines, as well as clear benefits from bottlenecked training and intent-aware cue extraction.

Our contributions include: 
(1) We introduce \textbf{InViC}, a lightweight and plug-in cue-based interface that distills dense visual tokens into compact, question-conditioned visual cues for Med-VQA.
(2) We propose a two-stage cue-bottleneck training strategy that explicitly constrains visual evidence flow to mitigate shortcut answering.
(3) We conduct extensive experiments on three public Med-VQA benchmarks and multiple MLLMs, demonstrating consistent gains and validating the importance of bottlenecked visual evidence flow.

\section{Methods}
\label{sec:method}

\subsection{Overview}
\label{subsec:overview}
\begin{figure}[t]
  \centering
  \includegraphics[width=\linewidth]{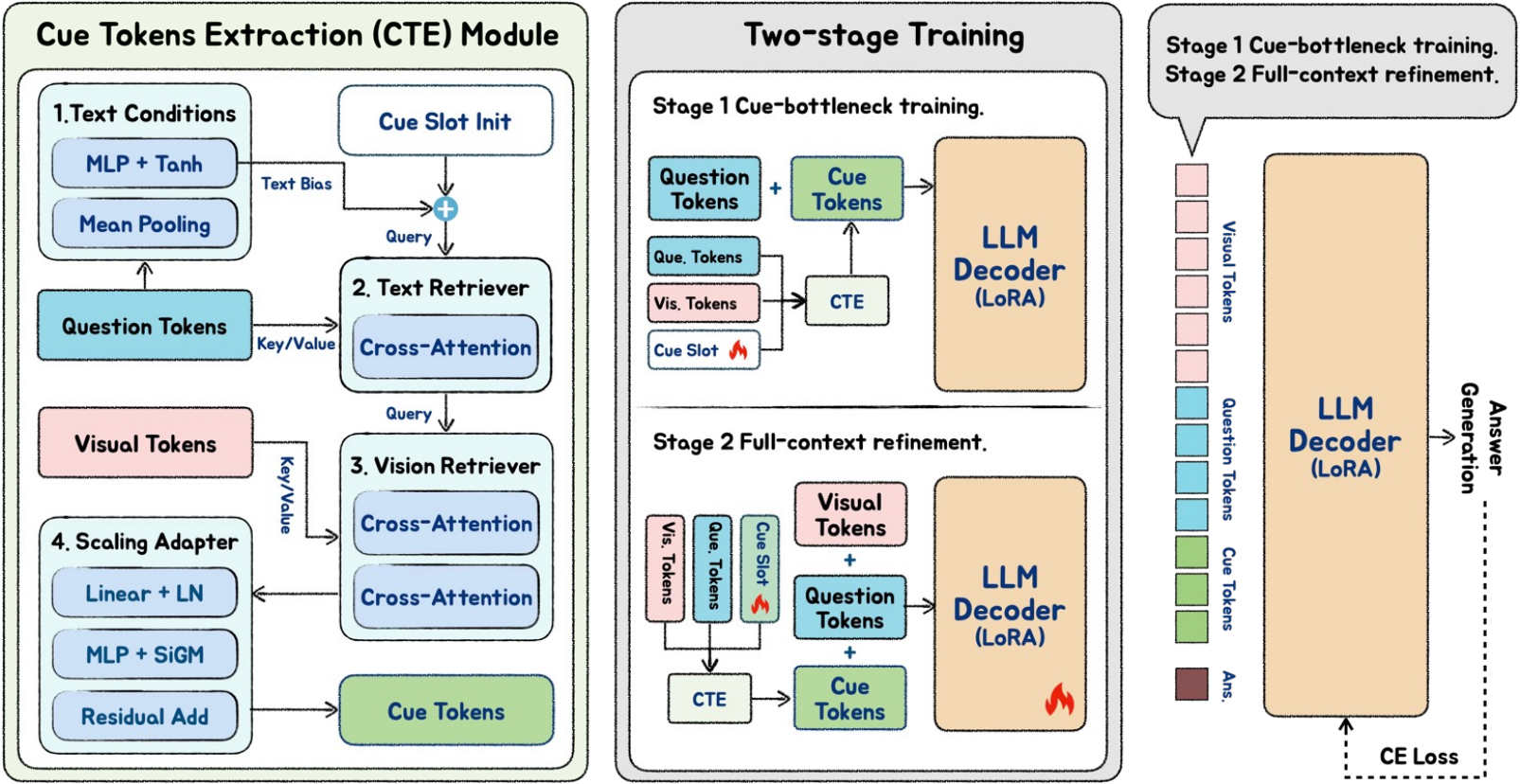}
  \caption{Overview of InViC. \textbf{Left:} CTE module with learnable cue slots initialized from a pooled question representation, followed by cross-attention to question tokens, cross-attention to visual tokens, and adapter-style calibration. \textbf{Middle:} Two-stage training strategy where visual tokens are visible only to cue tokens in Stage~I and fully visible in Stage~II. \textbf{Right:} Sequence construction and LLM training with cross-entropy loss.}
  \label{fig:overview}
\end{figure}

Given an image and a question, a vision encoder produces dense visual tokens and a text encoder produces question tokens. InViC inserts a CTE module between these backbone representations and the LLM decoder. The module generates $K$ question-guided cue tokens that summarize task-relevant visual evidence. These cue tokens are injected into the decoding sequence and participate in autoregressive answer generation.
To discourage shortcut answering, we adopt a two-stage fine-tuning strategy. Stage~I applies a cue-bottleneck attention mask to restrict direct access to raw visual tokens. Stage~II removes the bottleneck and refines the model under causal attention. The pipeline of InViC is shown in Fig.~\ref{fig:overview}.

\subsection{CTE Module}
\label{subsec:cue_module}
The cue extractor transforms backbone representations into $K$ cue tokens that are (i) semantically conditioned on the question intent and (ii) grounded in question-relevant image evidence. It consists of the following three steps.

\paragraph{Question-Conditioned Slot Initialization.}
Let $\{q_i\}_{i=1}^{N_q}$ denote the question token embeddings. We initialize $K$ learnable seed slots $S_{\mathrm{seed}}\in\mathbb{R}^{K\times d}$ with an offset derived from a pooled question representation:
\begin{equation}
S_0 = S_{\mathrm{seed}} + \tanh(\gamma)\,\mathrm{MLP}\!\left(\mathrm{MeanPool}(\{q_i\})\right),
\end{equation}
where $d$ is the cue-slot hidden dimension, $\gamma$ is a learnable scalar controlling the offset magnitude, and the offset is broadcast across all slots. In our implementation, this offset projection is a single-layer MLP initialized to zero, so the slots start from $S_{\mathrm{seed}}$ and gradually become question-conditioned during training.

\paragraph{Cross-Modal Retriever.}
We refine the cue slots via sequential cross-attention: slots first attend to the question tokens and then to the visual tokens to extract question-relevant evidence. Let $Q$ and $V$ denote the stacked question and visual token embeddings. Using standard cross-attention blocks,
\begin{equation}
S_1 = \mathrm{XBlock}(S_0, Q), \qquad 
S_2 = \mathrm{XBlock}(\mathrm{XBlock}(S_1, V), V),
\end{equation}
where $\mathrm{XBlock}(\cdot,\cdot)$ denotes a cross-attention transformer block. We use one block for text retriever and two blocks for visual retriever.

\paragraph{Residual Cue Calibration with Scaled Gating.}
We apply an adapter-style residual update with token-wise gating to stabilize cue refinement. Given the aggregated slots $S_2$, we first project them to the LLM hidden space and apply LayerNorm:
\begin{equation}
C_0 = \mathrm{LN}(S_2 W_P),
\end{equation}
where $W_P$ is a linear projection and $\mathrm{LN}(\cdot)$ denotes LayerNorm. A small MLP produces residual updates $\Delta=\mathrm{MLP}(C_0)$, which are gated with a sigmoid and scaled by a $\tanh$-constrained scalar:
\begin{equation}
C = C_0 + \tanh(\alpha)\,\bigl(w \odot \Delta\bigr), \qquad w=\sigma(\Delta),
\end{equation}
where $\sigma(\cdot)$ is sigmoid and $\odot$ denotes element-wise multiplication (broadcast over channels). We use $C$ as the calibrated cue tokens injected into the decoder context.

\subsection{Two-Stage Training with Cue Bottleneck}
\label{subsec:training}
To reduce shortcut answering under strong language priors, we train InViC in two stages: a cue-bottleneck stage that enforces cue-mediated perception and a full-context refinement stage. The model and optimization remain the same across stages; only the attention mask changes.

\paragraph{Stage~I: Cue-Bottleneck Training.}
We enforce a cue-bottleneck mask such that raw visual tokens $V$ are visible only to cue tokens $C$. We construct the LLM input sequence as $[V;\,Q;\,C;\,A]$, where cues are inserted between the question and answer tokens. On top of the standard causal and padding masks, we add
\begin{equation}
M_{\mathrm{bn}}(i,j)=
\begin{cases}
-\infty, & i\notin C \ \wedge\ j\in V,\\
0, & \text{otherwise},
\end{cases}
\end{equation}
where $i$ and $j$ index query and key/value positions, respectively, and $M_{\mathrm{bn}}$ is added to attention logits before softmax. This mask blocks any non-cue token (i.e., $Q$ and $A$) from attending to $V$, so image evidence can reach the decoder only through the cue tokens.

\paragraph{Stage~II: Full-Context Refinement.}
We keep the same input sequence $[V;\,Q;\,C;\,A]$ but remove the bottleneck and continue training from the Stage~I checkpoint using the standard causal attention mask. This refinement reduces the train-test mismatch introduced by the hard constraint, improves generation quality, and permits useful interactions with raw visual tokens while retaining the cue pathway learned in Stage~I.

\subsection{Training Objective}
\label{subsec:objective}
We optimize the standard next-token cross-entropy loss over answer positions:
\begin{equation}
\mathcal{L} = -\frac{1}{|\Omega|}\sum_{t\in\Omega}\log p(y_t\mid X, y_{<t}),
\end{equation}
where $X=[V;\,Q;\,C;\,A]$ denotes the input sequence, $y_t$ is the ground-truth token at step $t$, and $\Omega$ denotes the set of answer-token indices (i.e., positions corresponding to $A$).

\begin{figure}[t]
  \centering
  \includegraphics[width=\linewidth]{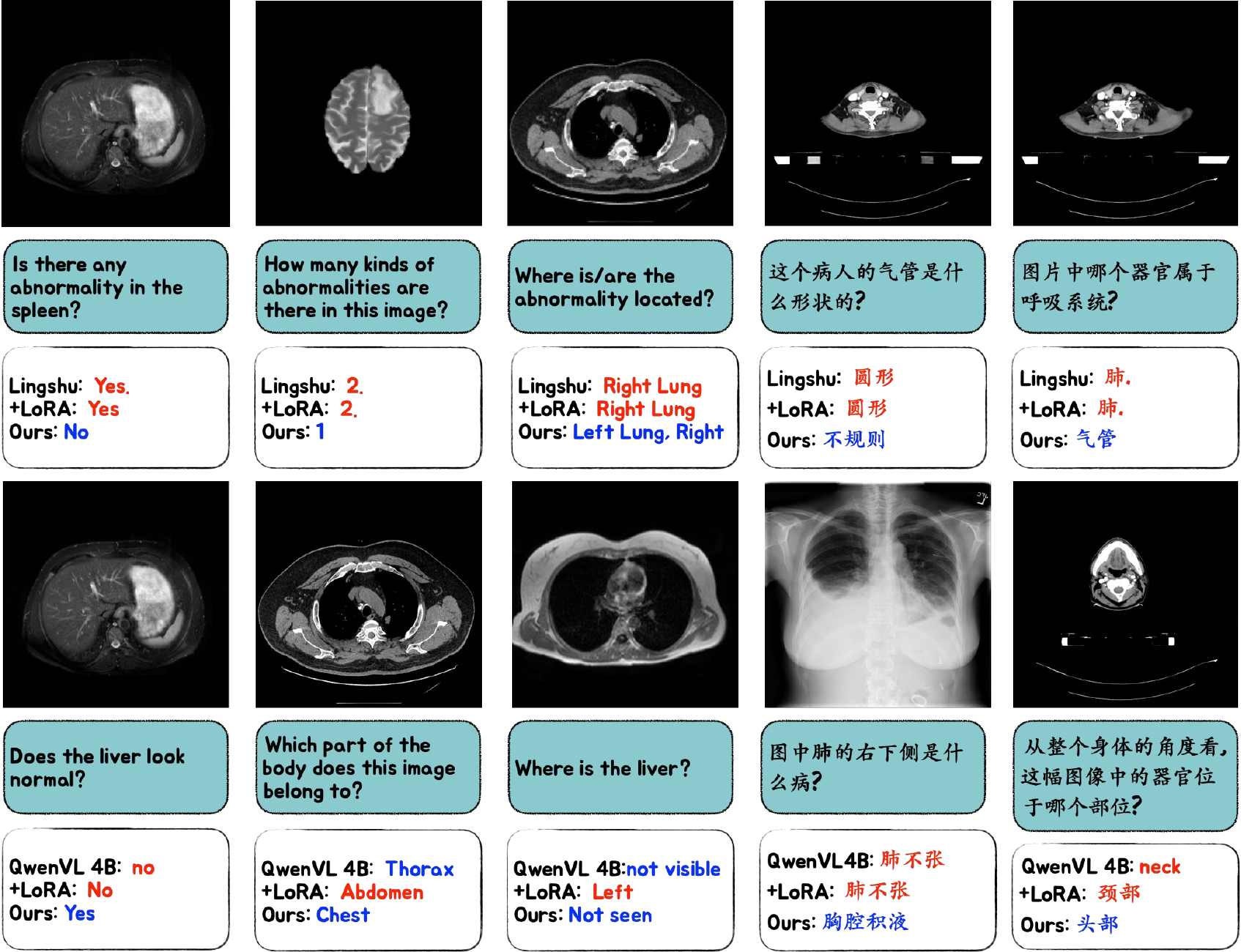}
  \caption{Qualitative comparison of Lingshu-7B and Qwen3-VL-4B on the SLAKE dataset. We visualize model predictions after zero-shot inference, LoRA fine-tuning, and integration of the InViC architecture. Blue highlights indicate correct predictions, while red indicates incorrect predictions.}
  \label{fig:result}
\end{figure}

\section{Experiments and Results}

\subsection{Datasets and Evaluation Metrics}

\subsubsection{Datasets.}
We evaluate InViC on three public Med-VQA benchmarks: VQA-RAD~\cite{lau2018dataset}, SLAKE~\cite{liu2021slake}, and ImageCLEF VQA-Med 2019~\cite{ben2019vqa}. 
VQA-RAD is a radiology-oriented dataset with clinically grounded questions that probe anatomy, imaging attributes, and abnormality-related concepts, with 1{,}793 training and 451 test question-answer pairs. 
SLAKE covers multiple imaging modalities (e.g., CT, MRI, and X-ray) and provides bilingual question-answer annotations in English and Chinese, enabling evaluation under mixed-language supervision, with 9{,}834 training, 2{,}099 validation, and 2{,}094 test pairs. 
ImageCLEF VQA-Med 2019 is a widely used benchmark for open-ended medical VQA, where answers are typically short textual phrases, with 12{,}792 training, 2{,}000 validation, and 500 test samples.

\subsubsection{Evaluation Metrics.}
We report accuracy using the unified evaluation rules of MedEvalKit\footnote{https://github.com/alibaba-damo-academy/MedEvalKit}. For closed-ended questions, we use exact match. For yes/no questions, we normalize the polarity. For open-ended questions, we employ an LLM-as-a-judge protocol with GPT-4o-mini to score correctness from (question, reference, prediction) triplets. We additionally report text-overlap metrics (Recall and BLEU) to characterize lexical similarity. Identical prompting and normalization are applied for all models to ensure fairness.

\subsection{Implementation Details}
We use $K{=}16$ cue tokens and set the LoRA rank to $r{=}8$. The cue extractor contains one text cross-attention block, two visual cross-attention blocks, and a residual cue calibration module. All experiments are conducted on a server with 8 NVIDIA GeForce RTX 4090 GPUs using AdamW, with a batch size of 8.
Training follows a two-stage schedule. In Stage~I, only the cue extractor is trained for 2 epochs with a learning rate of $1\times10^{-5}$ under the cue-bottleneck mask, while the backbone and LoRA adapters remain frozen. In Stage~II, the cue extractor and LoRA adapters are jointly trained for 3 epochs with a learning rate of $5\times10^{-6}$ under the standard causal mask, while the backbone remains frozen.

\begin{table}[t]
  \centering
  \footnotesize
  \caption{Main results on three Med-VQA benchmarks (VQA-Med 2019, SLAKE, and VQA-RAD). We report Accuracy, Recall, and BLEU for each method. In the \emph{LoRA fine-tuning and InViC integration} block, we highlight InViC results in bold.}
  \label{tab:main_results}
  \setlength{\tabcolsep}{1.5mm}
  \resizebox{\textwidth}{!}{%
  \begin{tabular}{lccc|ccc|ccc}
    \toprule
    \multirow{2}{*}{Method} & \multicolumn{3}{c|}{VQA-Med 2019} & \multicolumn{3}{c|}{SLAKE} & \multicolumn{3}{c}{VQA-RAD} \\
    \cline{2-10}
    & Accuracy & Recall & BLEU & Accuracy & Recall & BLEU & Accuracy & Recall & BLEU \\
    \midrule
    \multicolumn{10}{l}{\textbf{Closed-source MLLMs}} \\
    GPT-5.2 & 0.672 & 0.412 & 0.370 & 0.778 & 0.503 & 0.453 & 0.705 & 0.640 & 0.559 \\
    GPT-4.1 & 0.664 & 0.413 & 0.326 & 0.709 & 0.387 & 0.350 & 0.641 & 0.610 & 0.554 \\
    \addlinespace[0.3em]
    \multicolumn{10}{l}{\textbf{Medical-focused backbones}} \\
    Hulu-Med-4B~\cite{jiang2025hulumedtransparentgeneralistmodel} & 0.600 & 0.349 & 0.341 & 0.658 & 0.474 & 0.471 & 0.687 & 0.630 & 0.629 \\
    Hulu-Med-7B~\cite{jiang2025hulumedtransparentgeneralistmodel} & 0.636 & 0.362 & 0.354 & 0.702 & 0.511 & 0.510 & 0.747 & 0.686 & 0.688 \\
    Lingshu-7B~\cite{xu2025lingshu} & 0.606 & 0.397 & 0.386 & 0.840 & 0.804 & 0.805 & 0.678 & 0.608 & 0.607 \\
    \addlinespace[0.3em]
    \multicolumn{10}{l}{\textbf{General vision-language backbones}} \\
    Qwen3-VL-4B~\cite{Qwen3-VL} & 0.572 & 0.306 & 0.294 & 0.592 & 0.432 & 0.432 & 0.599 & 0.535 & 0.538 \\
    Qwen3-VL-8B~\cite{Qwen3-VL} & 0.540 & 0.342 & 0.336 & 0.569 & 0.433 & 0.433 & 0.648 & 0.559 & 0.558 \\
    InternVL3.5-4B~\cite{wang2025internvl3} & 0.642 & 0.519 & 0.527 & 0.697 & 0.627 & 0.622 & 0.656 & 0.604 & 0.604 \\
    InternVL3.5-8B~\cite{wang2025internvl3} & 0.666 & 0.592 & 0.599 & 0.678 & 0.599 & 0.599 & 0.670 & 0.606 & 0.607 \\
    \midrule
    \multicolumn{10}{l}{\textbf{LoRA fine-tuning and InViC integration}} \\
    Lingshu-7B + LoRA & 0.654 & 0.642 & 0.645 & 0.845 & 0.809 & 0.809 & 0.680 & 0.611 & 0.609 \\
    Lingshu-7B + InViC & \textbf{0.672} & \textbf{0.667} & \textbf{0.668} & \textbf{0.868} & \textbf{0.840} & \textbf{0.841} & \textbf{0.688} & \textbf{0.625} & \textbf{0.624} \\
    Qwen3-VL-4B + LoRA & 0.654 & 0.633 & 0.633 & 0.826 & 0.806 & 0.808 & 0.612 & 0.566 & 0.571 \\
    Qwen3-VL-4B + InViC & \textbf{0.697} & \textbf{0.627} & \textbf{0.622} & \textbf{0.849} & \textbf{0.832} & \textbf{0.835} & \textbf{0.632} & \textbf{0.571} & \textbf{0.576} \\
    Qwen3-VL-8B + LoRA & 0.650 & 0.639 & 0.644 & 0.820 & 0.803 & 0.808 & 0.665 & 0.599 & 0.601 \\
    Qwen3-VL-8B + InViC & \textbf{0.676} & \textbf{0.655} & \textbf{0.656} & \textbf{0.835} & \textbf{0.810} & \textbf{0.812} & \textbf{0.674} & \textbf{0.614} & \textbf{0.615} \\
    \bottomrule
  \end{tabular}%
  }
\end{table}

\subsection{Results}

\subsubsection{Comparison with Representative MLLMs.}
We compare InViC with representative MLLMs, including closed-source models (GPT-5.2 and GPT-4.1) and open-source backbones under zero-shot inference and LoRA fine-tuning. As summarized in Table~\ref{tab:main_results}, InViC consistently improves accuracy across all three benchmarks. For example, on Qwen3-VL-4B, InViC improves accuracy on SLAKE from 0.592 (zero-shot) to 0.849, and also surpasses LoRA (0.826 to 0.849); on VQA-Med 2019, it improves from 0.572 to 0.697. On Lingshu-7B, InViC further improves accuracy on SLAKE over LoRA from 0.845 to 0.868. Figure~\ref{fig:result} presents five question-answer cases for each of Lingshu-7B and Qwen3-VL-4B on SLAKE, illustrating that InViC yields more accurate predictions by leveraging question-conditioned cues.

\subsubsection{Ablation Studies.}
We conduct ablations on SLAKE with Qwen3-VL-4B to study Stage~I cue-bottleneck training, the cue design, and the cue token count $K$. Here, learnable tokens denote $K$ randomly initialized embeddings inserted at the cue positions without CTE. As shown in Table~\ref{tab:ablation}, Stage~II-only learnable tokens perform poorly (accuracy 0.609), while adding Stage~I improves accuracy to 0.799. Replacing learnable tokens with cue tokens further improves accuracy to 0.849 under Stage~I+II training, suggesting the benefit of CTE over generic inserted token embeddings. Similar trends are observed for recall and BLEU.
We further vary $K\in\{4,8,16,32\}$ and find that performance peaks at $K{=}16$, with $K{=}8$ being competitive, while too few cue tokens limit capacity and too many introduce redundancy.

\begin{table}[t]
\centering
\footnotesize
\setlength{\tabcolsep}{1.1mm}
\caption{Ablation results on SLAKE with Qwen3-VL-4B. \textbf{Left:} comparison between learnable tokens and cue tokens under different training schedules ($K{=}16$). \textbf{Right:} effect of the cue token count $K$.}
\label{tab:ablation}
\begin{tabular}{lcc|cc|cccc}
\toprule
\multicolumn{1}{c}{} & \multicolumn{2}{c|}{Learnable tokens} & \multicolumn{2}{c|}{Cue tokens} & \multicolumn{4}{c}{Cue token count} \\
\cmidrule(r){2-5}\cmidrule(l){6-9}
Metric & \makecell{Stage~II} & \makecell{Stage~I$+$II} & \makecell{Stage~II} & \makecell{Stage~I$+$II} & $K{=}4$ & $K{=}8$ & $K{=}16$ & $K{=}32$ \\
\midrule
Accuracy & 0.609 & 0.799 & 0.770 & 0.849 & 0.785 & 0.832 & 0.849 & 0.800 \\
Recall & 0.448 & 0.745 & 0.699 & 0.832 & 0.733 & 0.809 & 0.832 & 0.740 \\
BLEU & 0.449 & 0.747 & 0.701 & 0.835 & 0.735 & 0.811 & 0.835 & 0.740 \\
\bottomrule
\end{tabular}
\end{table}

\section{Conclusion}
We proposed \textbf{InViC}, an intent-aware visual cue interface for Med-VQA that distills dense visual tokens into a compact set of question-guided cue tokens and injects them into the LLM decoding context. 
To mitigate shortcut answering, we introduced a two-stage fine-tuning recipe with a cue-bottleneck attention mask that enforces visual information flow through the cue pathway in Stage~I and restores standard causal attention in Stage~II. Experiments on three public Med-VQA benchmarks demonstrate consistent gains over zero-shot inference and standard LoRA fine-tuning across representative MLLM backbones. 
Ablations further validate the importance of Stage~I bottleneck training and the cue token count. 
Future work will extend InViC to additional backbones and clinical datasets and explore more interpretable cue designs.

\bibliographystyle{splncs04}
\bibliography{refer}

\end{document}